\documentclass[aps,preprint,showpacs,preprintnumbers,amsmath,amssymb,superscriptaddress]{revtex4-1}
\usepackage{graphics}
\usepackage[pdftex]{graphicx}
\usepackage{color}
\usepackage{multirow}

\begin{document}

\title{ Scaling laws in human speech, decreasing emergence of new words and a generalized model}

\author{Ruokuang Lin}
\affiliation{\small School of Electronic Science and Engineering, Ministry of Education Key Laboratory of Modern Acoustics, Institute for Biomedical Electronics Engineering, Nanjing University, Nanjing 210093, China}

\author{Chunhua Bian}
\email[Corresponding author:]{bch@nju.edu.cn}
\affiliation{\small School of Electronic Science and Engineering, Ministry of Education Key Laboratory of Modern Acoustics, Institute for Biomedical Electronics Engineering, Nanjing University, Nanjing 210093, China}
\affiliation{\small Department of Physics, Boston University, Boston, Massachusetts 02215, USA}
\affiliation{\small Department of Neurology, Beth Israel Deaconess Medical Center and Harvard Medical School, Boston, Massachusetts 02215, USA}

\author{Qianli D. Y. Ma}
\email[Corresponding author:]{maql@njupt.edu.cn}
\affiliation{\small Department of Physics, Boston University, Boston, Massachusetts 02215, USA}
\affiliation{\small College of Geographic and Biologic Information, Nanjing University of Posts and Telecommunications, Nanjing 210003, China}

\begin{abstract}

\noindent Human language, as a typical complex system, its organization and evolution is an attractive topic for both physical and cultural researchers. In this paper, we present the first exhaustive analysis of the text organization of human speech. Two important results are that: (i) the construction and organization of spoken language can be characterized as Zipf's law and Heaps' law, as observed in written texts; (ii) word frequency vs. rank distribution and the growth of distinct words with the increase of text length shows significant differences between book and speech. In speech word frequency distribution are more concentrated on higher frequency words, and the emergence of new words decreases much rapidly when the content length grows. Based on these observations, a new generalized model is proposed to explain these complex dynamical behaviors and the differences between speech and book. 

\medspace
\medspace


\end{abstract}

\maketitle

\section*{Introduction}

Numerous statistic studies have been done to uncover the dynamical universal laws in complex system in physical, biological and social areas, such as the reaction dynamics within cells, financial market fluctuations, income distribution, biological species, word frequency, scientific publication, city size, etc.[1-19]. Zipf's law and Heaps' law are two typical representatives, the coexistence of which have been observed in various systems [20, 21]. Zipf finds a certain scaling law in the rank of the word frequencies. Heaps' law reveals another aspect of scaling in natural language processing, according to which the vocabulary size grows in a sub-linear function with document size. It is also found that Zipf's law and Heaps' law holds for different languages with different characterization exponents which means more complicated statistical features [22-27]. Moreover, similar results have been recently reported for the corpus of web texts, including collaborative tagging, social annotation, internet search result and computer programs[28-32], which indicates universality of Zipf's law and Heaps' law beyond natural language systems.

In this paper, we focus on the word frequency in spoken language. Using the speech transcriptions from the Speech British National Corpus on daily conversations of various subjects and ten classic books for comparison, we show (i) the word frequency distribution vs. ranking obeys the Zipf's law and growing of distinct words obeys Heaps' law; (ii) the coefficients of Zipf's and Heaps' laws have significant differences between written texts and speech transcriptions; and (iii) on the basis of the above study, a generalized model is proposed to simulate the growing dynamics and construction mechanism of spoken and written languages and the difference between speech and written texts. Empirical observations and model simulations agree well with each other.

\section*{Results}

\noindent \textbf{Experiments.} Two classes of data sets were analyzed in this article: (i) Ten speech transcriptions of the Speech British National Corpus [33] from Phonetics Laboratory, and (ii) ten classic books written by different authors in different era. The transcriptions of daily conversation from BNC database are selected to have comparable length as the books. As shown in Table 1, the total lengths of the text are 27341 to 268843 for books, and 25281 to 124802 for speeches. There are relatively less distinct words being used in speeches, ranging from 2158 to 5815 in the selected data sets, and from 2572 to 29020 in books (see details about data in Methods).

Firstly, we analyzed the probability distribution of word frequency. We analyzed each book and speech transcription. The probability distribution of word frequency can be described as a power-law between the word frequency $k$ and its probability density $P(k)$, 

\begin{equation}
P(k) \sim k^{-\beta}, \label{M1}
\end{equation}

\noindent where $\beta$ is the power-law scaling exponent. Fig. 1 shows the probability distribution of word frequency $P(k)$ of two sets of data, the result of book is in Fig. 1(a) and speech is in Fig. 1(b), which both follow power-law. The goodness of fit is 0.9952 and 0.9944 respectively, and fitting region is $k=2 \sim 100$ (see details of the goodness of fit in Methods). All the validation results are listed in Table S1. We found significant difference of scaling exponents between books and speeches (book: $1.77 \pm 0.11$, speech: $1.67 \pm 0.05$, $p<0.05$). 

\begin{figure}
\centerline{
\includegraphics[width=0.5\linewidth]{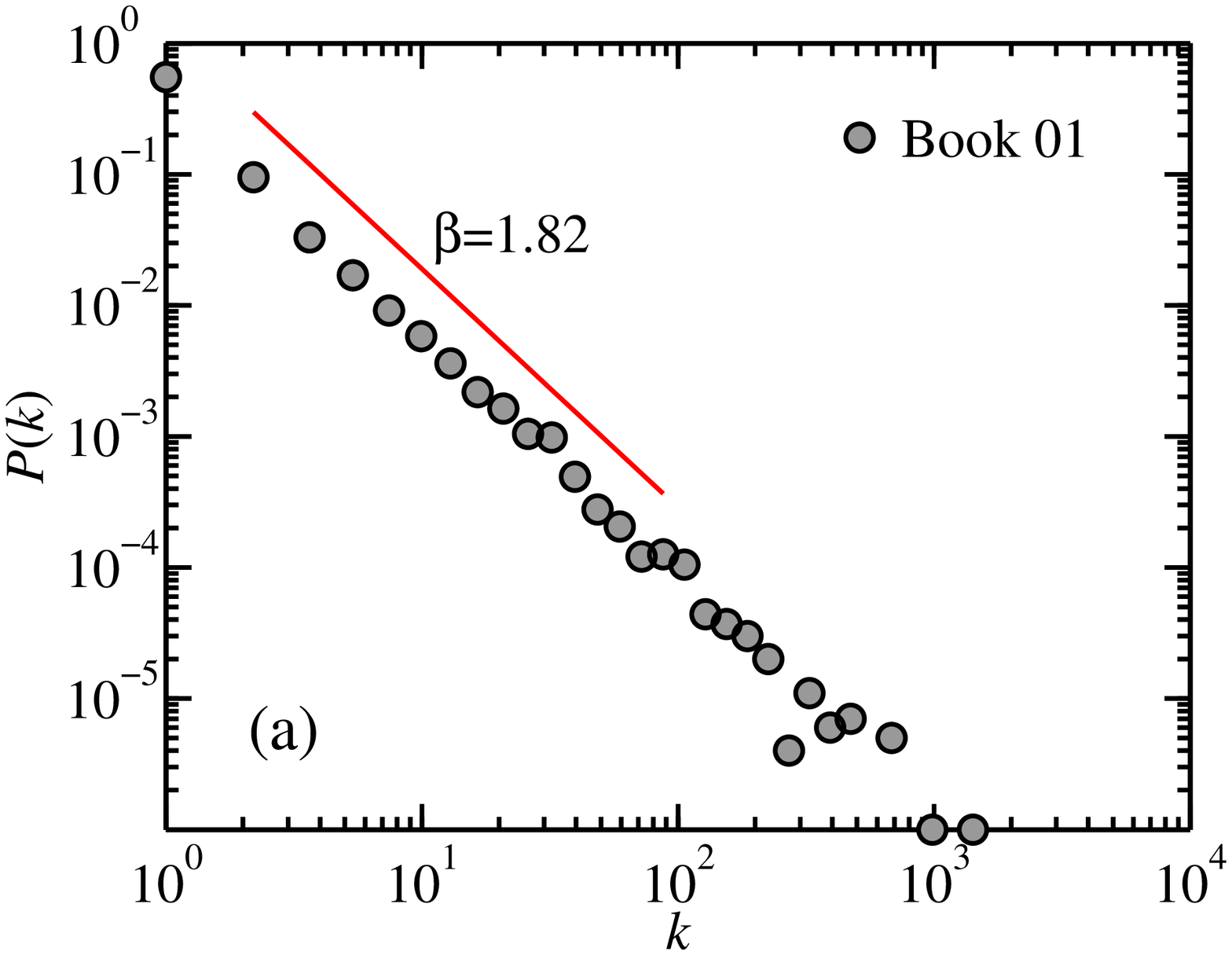}
\includegraphics[width=0.5\linewidth]{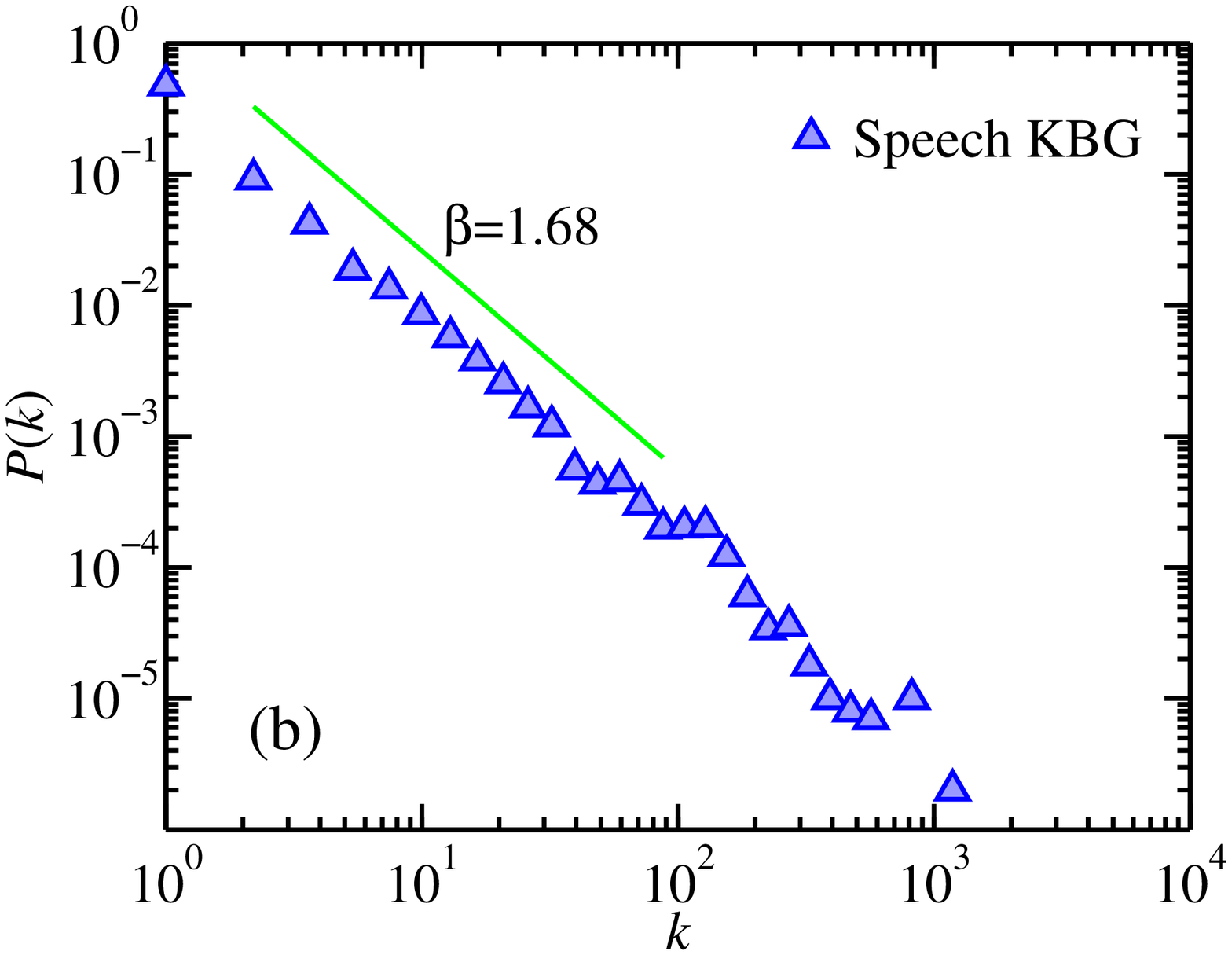}
}
\caption{The probability distribution of word frequency $P(k)$ vs. word frequency $k$ in log-log scale for (a) books and (b) speech transcriptions. The fitting regions are $k=2 \sim 100$.}
\end{figure}

Zipf found a power-law relation between the work frequency $Z(r)$, and its corresponding rank $r$, as

\begin{equation}
Z(r) \sim r^{-\alpha}. \label{M2}
\end{equation}

Fig. 2(a) presents the plot of word frequency distribution $Z(r)$ for one book and one speech transcription. The word frequency distributions can be divided into two parts, in the first part, which is corresponding to the high frequency words, the relative word frequency $Z(r)$ of books is less than that of speech. In the second part, which is corresponding to the low frequency words, the $Z(r)$ of books is larger than that of speech. The decay of the second part for speech and book both follow power-law with the goodness of fit 0.9985 and 0.9992 respectively and we found significant difference in the decay exponents $\alpha$ between books and speeches (book: $1.12 \pm 0.08$, speech: $1.39 \pm 0.06$, $p<0.01$, see Table S1).

\begin{figure}
\centerline{
\includegraphics[width=0.5\linewidth]{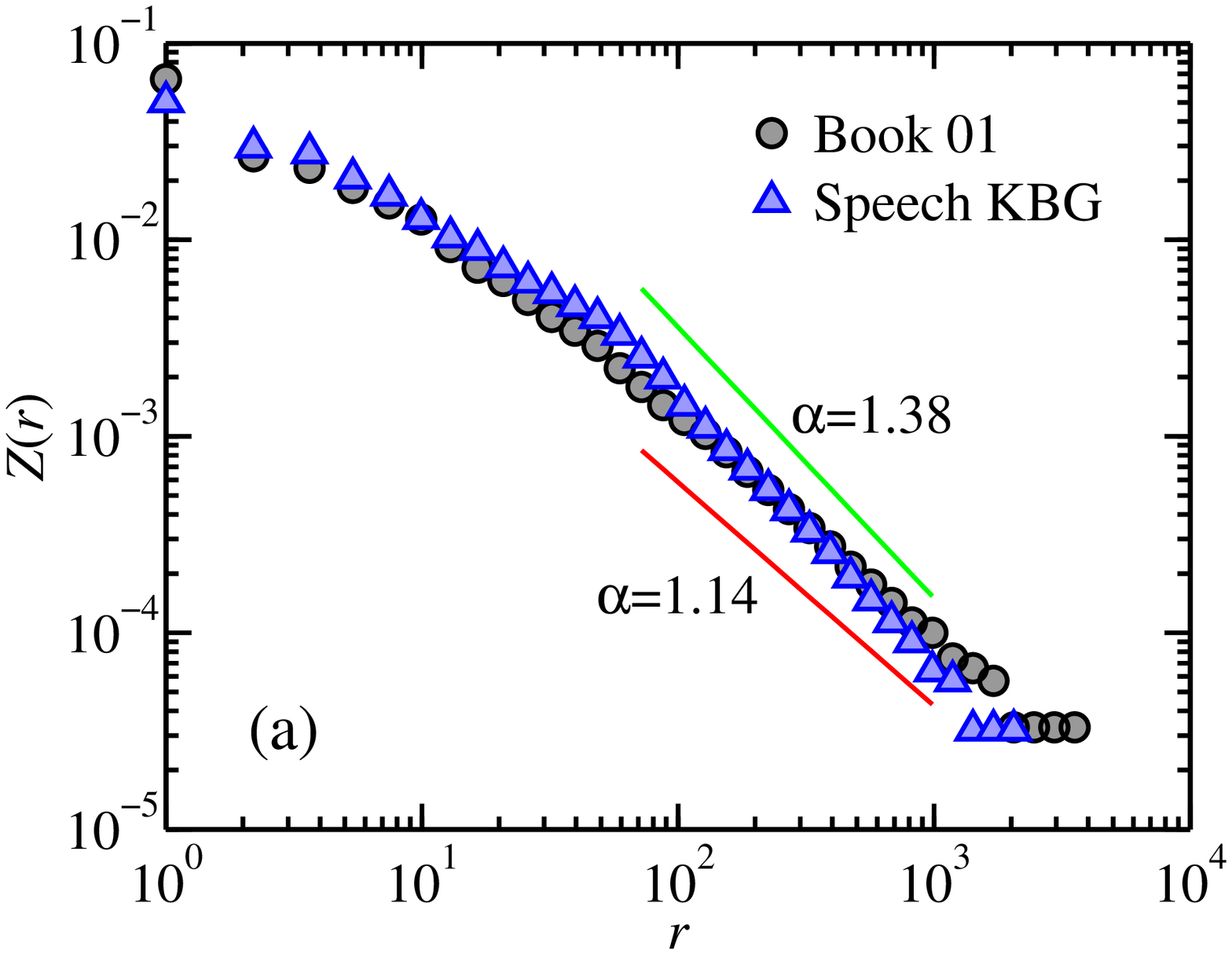}
\includegraphics[width=0.5\linewidth]{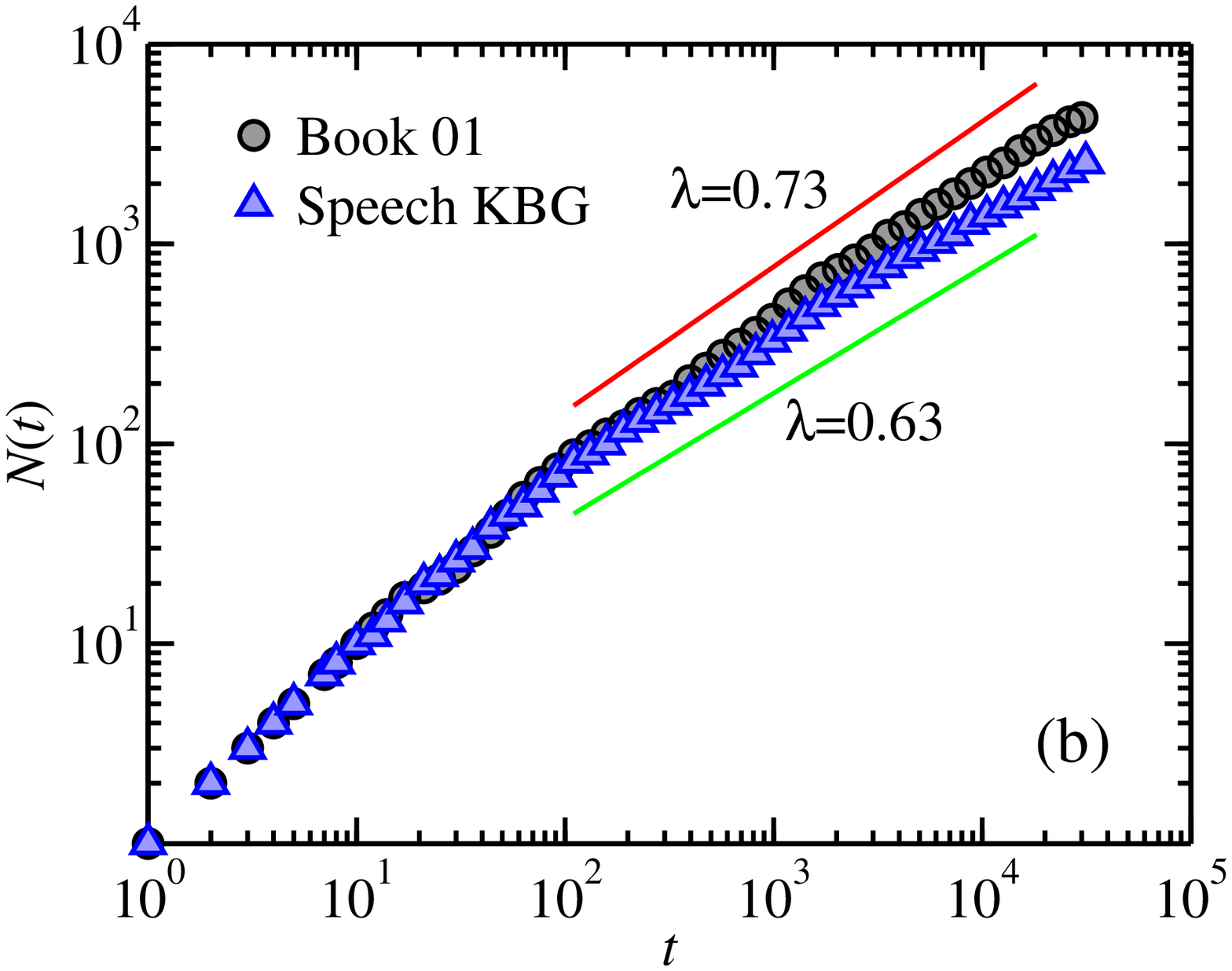}
}
\caption{(a) The Zipf's plot on word frequency $Z(r)$ of one book and one speech transcription, with a fitting region $r=60 \sim 1000$. (b) Comparison of the growth of the number of distinct words $N(t)$ versus the text length $t$ between one book and one speech transcription, fitting region $t=100 \sim 20000$.}
\end{figure}

Fig. 2(b) reports the growth of the number of distinct words N(t) versus the text length t. Both books and speeches follow Heaps' law

\begin{equation}
N(t) \sim t^{\lambda}, \label{M3}
\end{equation}

\noindent and can also be divided into two parts. In the first part, $N(t)$ of the speeches is very close to that of the books, corresponding to the linear increasing region. In the second part, the sub-linear increasing region, we can see that $N(t)$ of the books is bigger than those of the speeches, which indicates that the growth speed of new words in speech is lower than in book.

We calculated the slope $\lambda$ of the second part of $t \sim N(t)$ curve of ten books and ten speech transcriptions as a linear approximate fitting in log-log scale, and the slope values of each subject were listed in Table 1. The $t$-test shows there is a significant difference between the slope of books and speeches (book: $0.73 \pm 0.04$, speech: $0.63 \pm 0.03$, $p<0.01$), which both follow power-law with the goodness of fit is 0.9988 and 0.9980 respectively, and fitting region is $k=100 \sim 20000$.

\begin{table}[ht]
\caption{The basic statistics of the ten speech transcriptions and the ten books. $T$ is the total length of the text measured in number of words, and $Nt$ is the total number of distinct words. The slope of $t \sim N(t)$ of the ten speech transcriptions and the ten books (local linear approximate fitting in log-log scale, fitting region $t=100 \sim 20000$)}
\bigskip

\begin{tabular}{cccccccc}
\hline
\hline
\multirow{2}{*}{~~No.~~} & \multicolumn{3}{c}{Book} & & \multicolumn{3}{c}{Speech} \\
\cline{2-4} \cline{6-8}
& $T$ & $Nt$ & $\lambda$ & &  $T$ & $Nt$ & $\lambda$ \\
\hline
01 & 30083 & 4285 & 0.73 & & 112806 & 4729 & 0.60 \\
02 & 27341 & 2572 & 0.65 & & 113067 & 4978 & 0.62 \\
03 & 57314 & 4778 & 0.72 & & 31349 & 2551 & 0.63 \\
04 & 78802 & 7343 & 0.76 & & 29934 & 2411 & 0.64 \\
05 & 30979 & 4834 & 0.75 & & 124802 & 4236 & 0.64 \\
06 & 218510 & 16963 & 0.79 & & 25281 & 2158 & 0.65 \\
07 & 73908 & 7172 & 0.73 & & 54349 & 2856 & 0.60 \\
08 & 59909 & 5546 & 0.72 & & 65965 & 3295 & 0.61 \\
09 & 156815 & 6930 & 0.67 & & 116725 & 5815 & 0.68 \\
10 & ~268843~ & ~29020~ & ~0.75~ &~~& ~73145~ & ~4488~ & ~0.66~ \\
\hline
% & ~$1.770 \pm 0.110$~ & $1.662 \pm 0.054$~ & ~$1.120 \pm 0.078$~ & $1.393 \pm 0.060$~ & ~$0.727 \pm 0.041$~ & $0.633 \pm 0.026$ \\
% & \multicolumn{2}{c}{$p=0.03$} & \multicolumn{2}{c}{$p<0.01$} & \multicolumn{2}{c}{$p<0.01$} \\
\hline
\end{tabular}
\end{table}

Fig. 3 are the summary of mean and standard deviation of the scaling exponents $\beta$, $\alpha$ and $\lambda$ of books and speeches.

\begin{figure}
\centerline{
\includegraphics[width=0.5\linewidth]{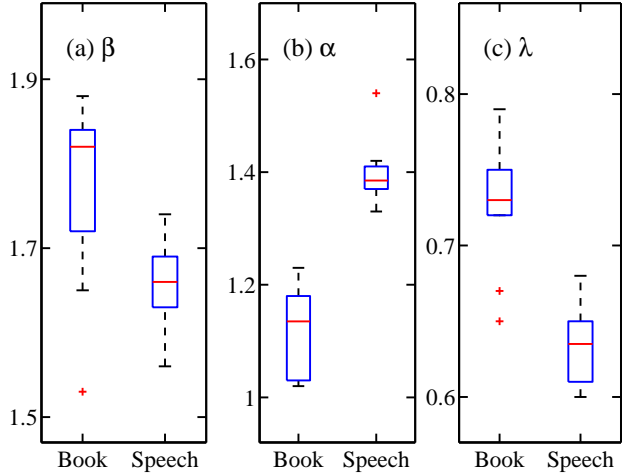}
}
\caption{The mean and standard deviation of scaling exponents for (a) the probability distribution of word frequency, (b) Zipf's law and (c) Heaps' law. Significant differences were found between speech and book.}
\end{figure}

\noindent \textbf{Model.} We test whether the rich-get-richer mechanism, also named preference attachment mechanism [34-37] works for spoken language generating process. We denote $\phi(k)$ the average probability that a character appeared $k$ times will appear again (see Methods how to measure $\phi(k)$). As shown in Fig. 4, $\phi(k)$ for all the books and speeches increase proportionally with $k$, indicating a rich-get-richer effect like the preferential attachment in evolving scale-free networks.

\begin{figure}
\centerline{
\includegraphics[width=0.5\linewidth]{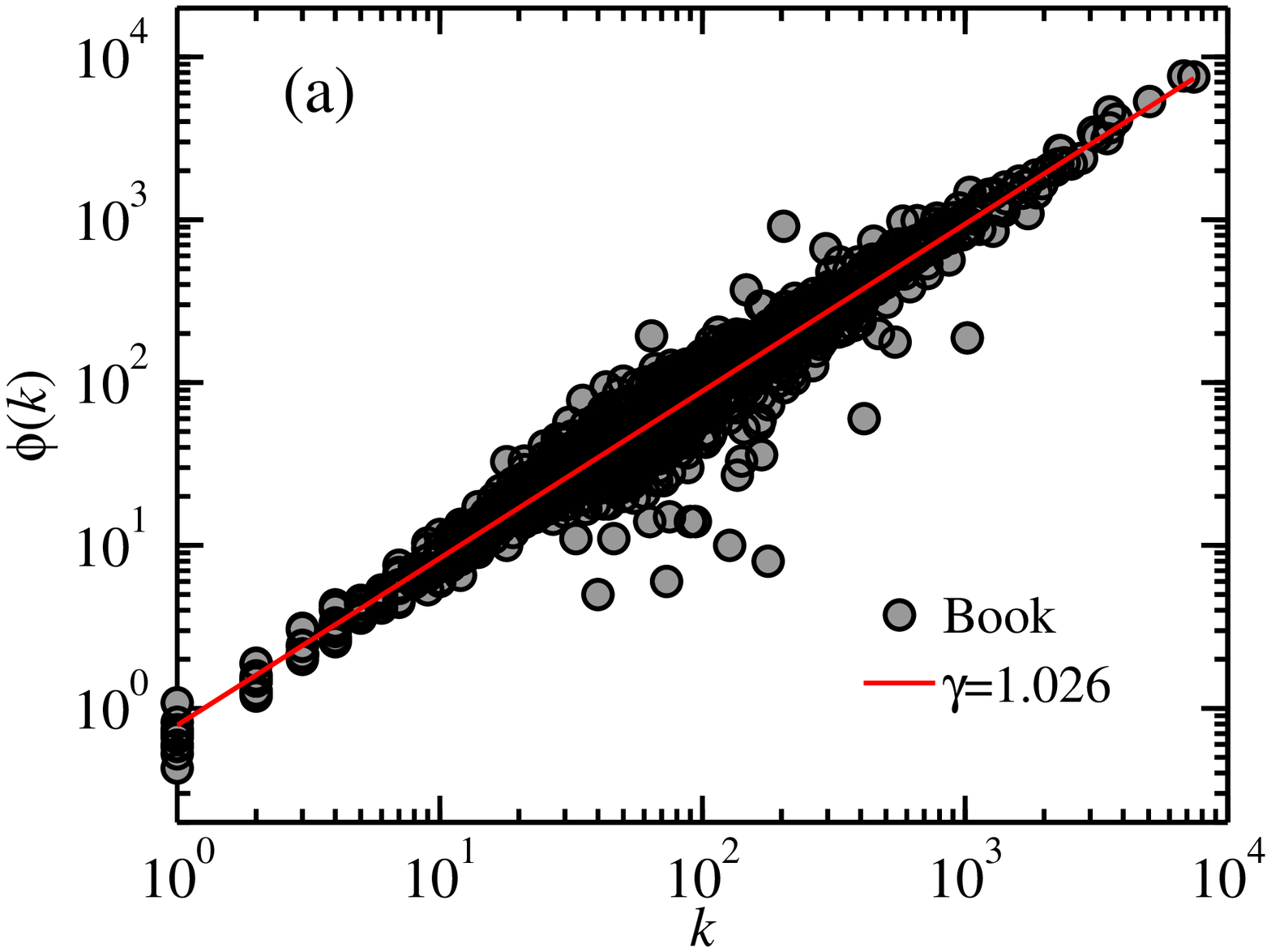}
\includegraphics[width=0.5\linewidth]{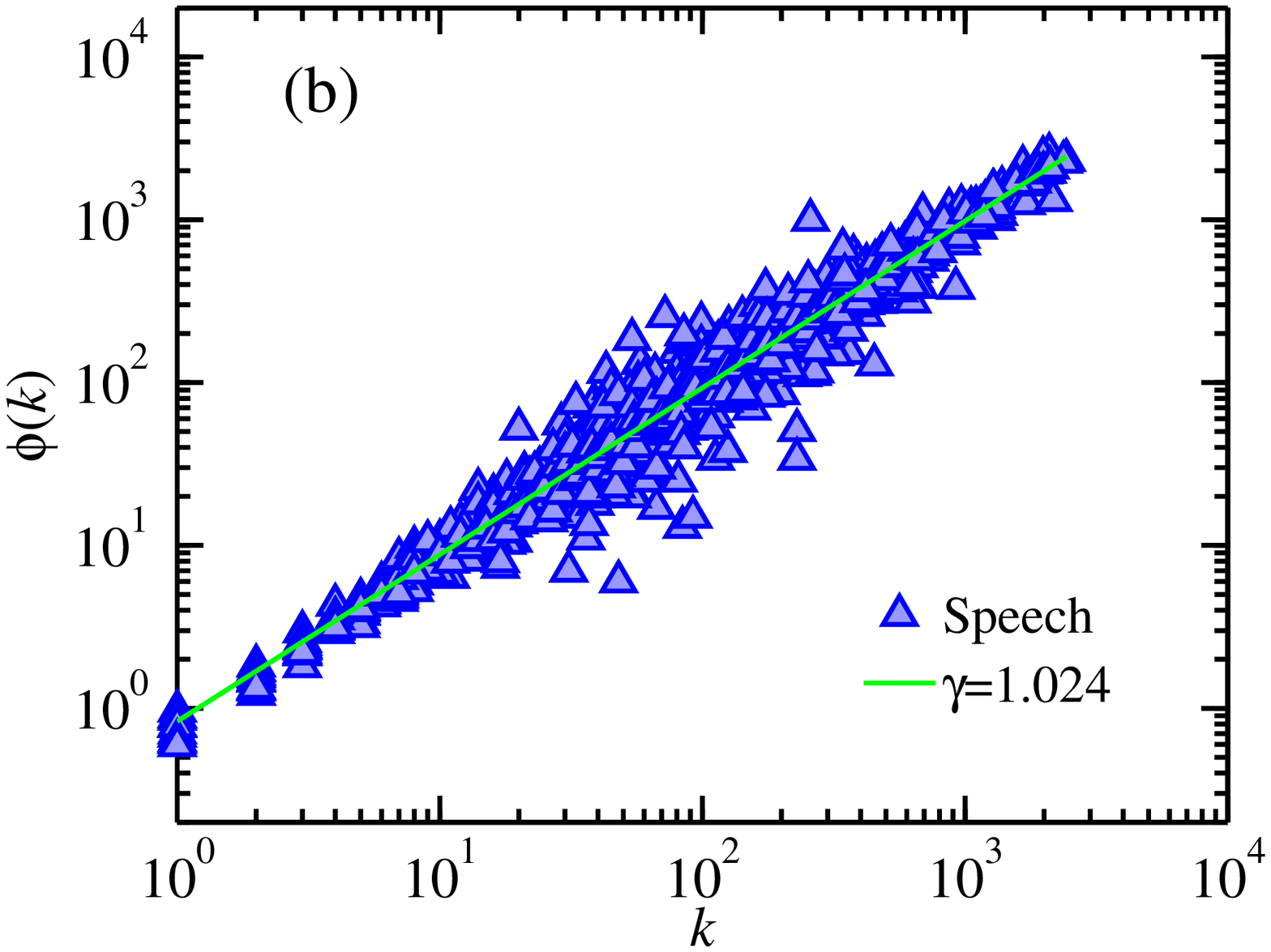}
}
\caption{$\phi(k)$ versus $k$ in a log-log scale for the representative (a) book and (b) speech, and the goodness of fit are 0.9477 and 0.9575 respectively.}
\end{figure}

We propose a generalized model to further simulate and investigate the empirical observations as an extensive Yule-Simon model [38-43]. The process of language construction can be modeled as follows. At every time step, a word will be appended to the text, either by generating a new word, or selecting one from the text that already generated. We propose that the growing dynamics revealed by Heaps' law can be defined as the probability $p$ of new word generation, which can gradually change with the text length. The fomula of the probability $p$ can be determined according to the specific application. In this investigation of language construction, we set it as:

\begin{equation}
p = k_0t^{k_t}. \label{equ:model1}
\end{equation}

While with probability $\overline{p}=1-p$, one word is copied from the text that already generated, where word will be chosen is determined by rich-get-richer mechanism. Let $n(i, t)$ be the number of appearance of $i$th word at time step $t$, then at next step $t+1$, the $i$th word will be selected with the probability

\begin{equation}
p(i, t+1) = (1-p){n(i, t)^{k_p} \over \sum_{i} n(i, t)^{k_p}}. \label{equ:model2}
\end{equation}

\noindent Parameter $k_p$ provides a parameter to modulate the strength of preferential attachment. In this proposed model, the probability $p$ of new word generation depends on text length $t$, and decreases monotonously with text increasing, inspired by the empirical observation of $t \sim N(t)$. And the word will be reused according to the rich-by-richer rules, which gives the Zipf's law. Fig. 5 reports the simulation results for one book and one speech transcription using proposed model. All three scaling properties can be very well captured by the model. For all the ten books and ten speech transcriptions, the parameters are as follows: $k_0$, book: $2.93 \pm 1.01$, speech: $3.26 \pm 0.72$, $p=0.5$; $k_t$, book: $0.33 \pm 0.05$, speech: $0.41 \pm 0.04$, $p<0.01$; $k_p$, book: $1.10 \pm 0.02$, speech: $1.06 \pm 0.01$, $p<0.01$ (see Table S2 and Fig. S1). 

\begin{figure}
\centerline{
\includegraphics[width=0.33\linewidth]{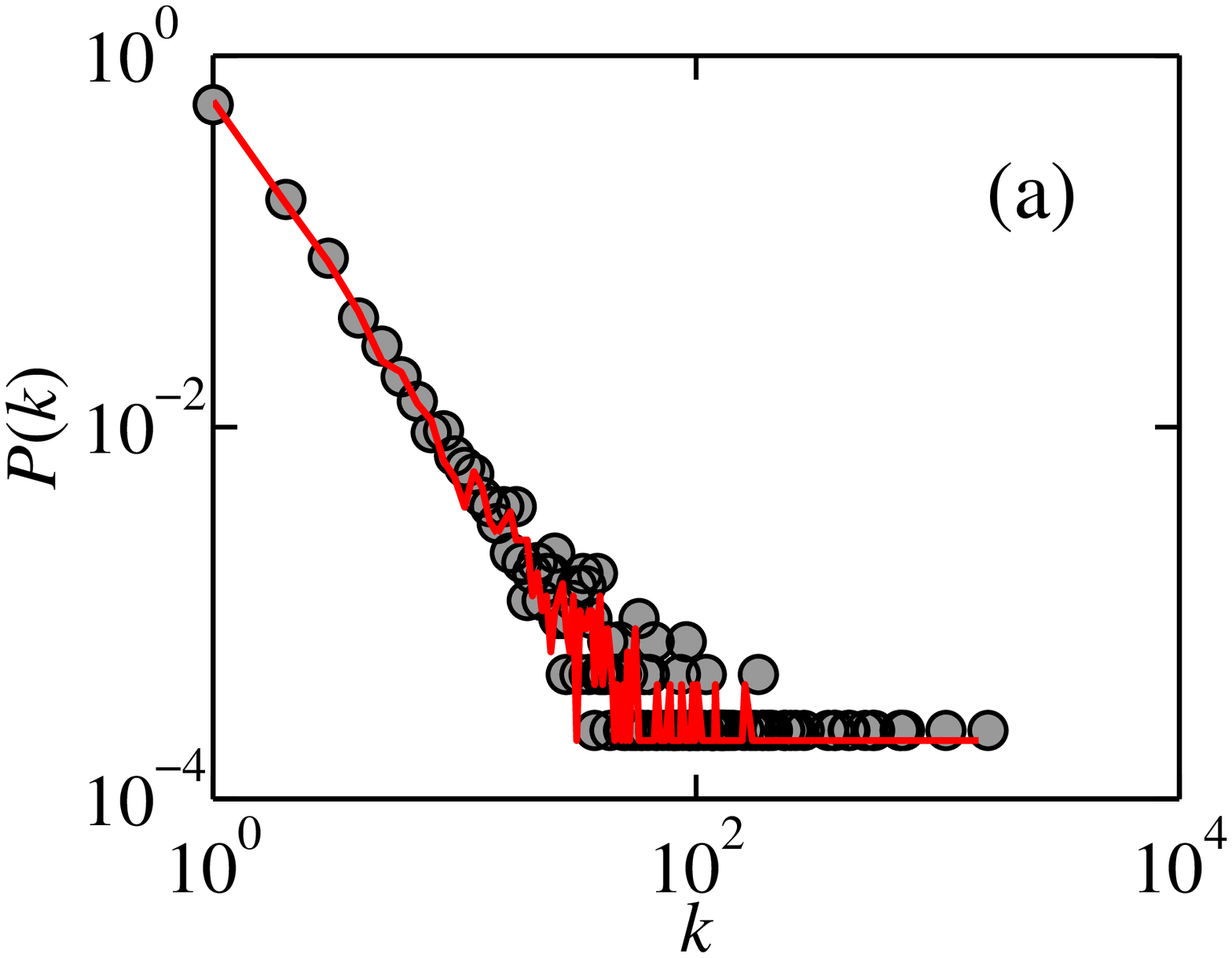}
\includegraphics[width=0.33\linewidth]{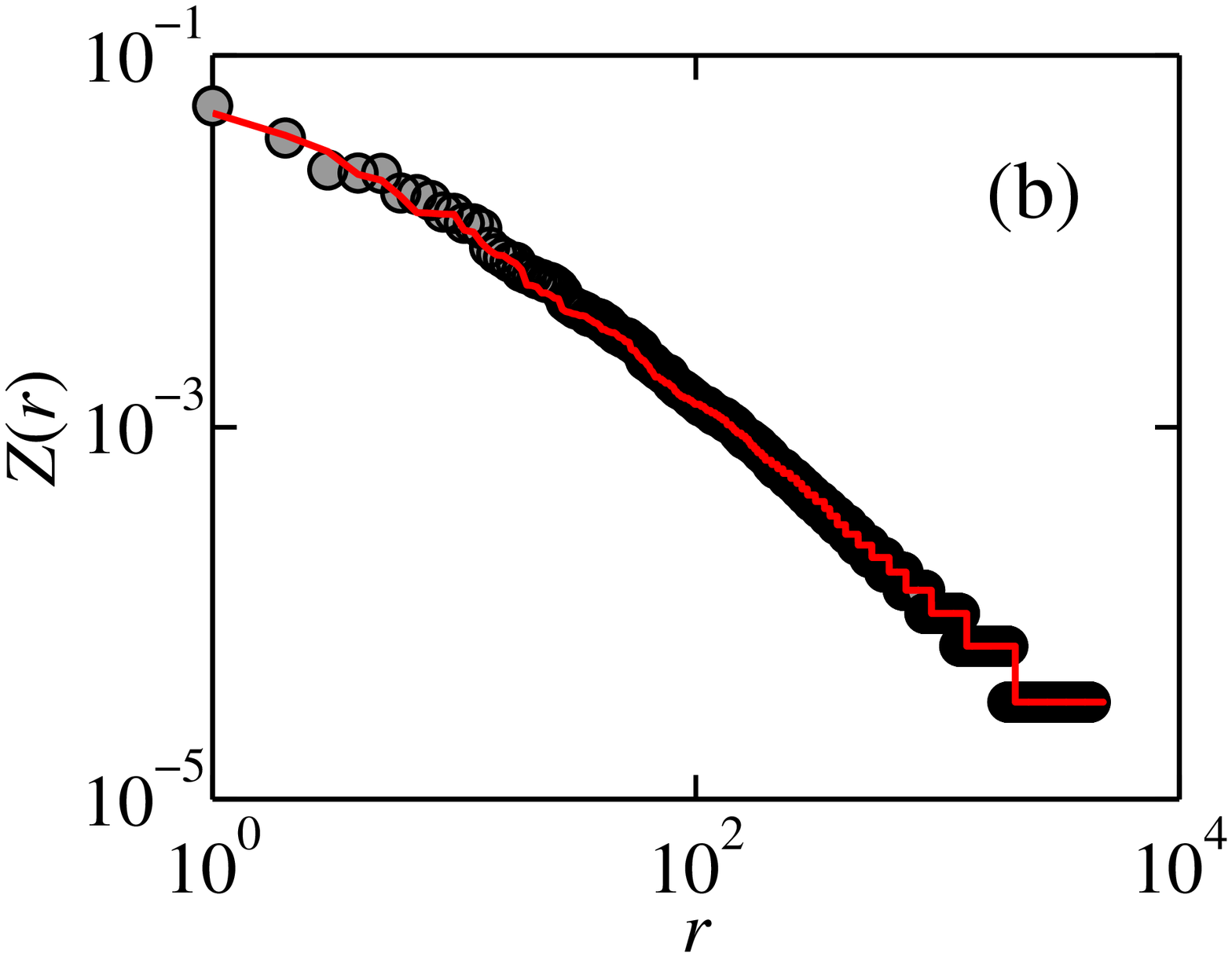}
\includegraphics[width=0.33\linewidth]{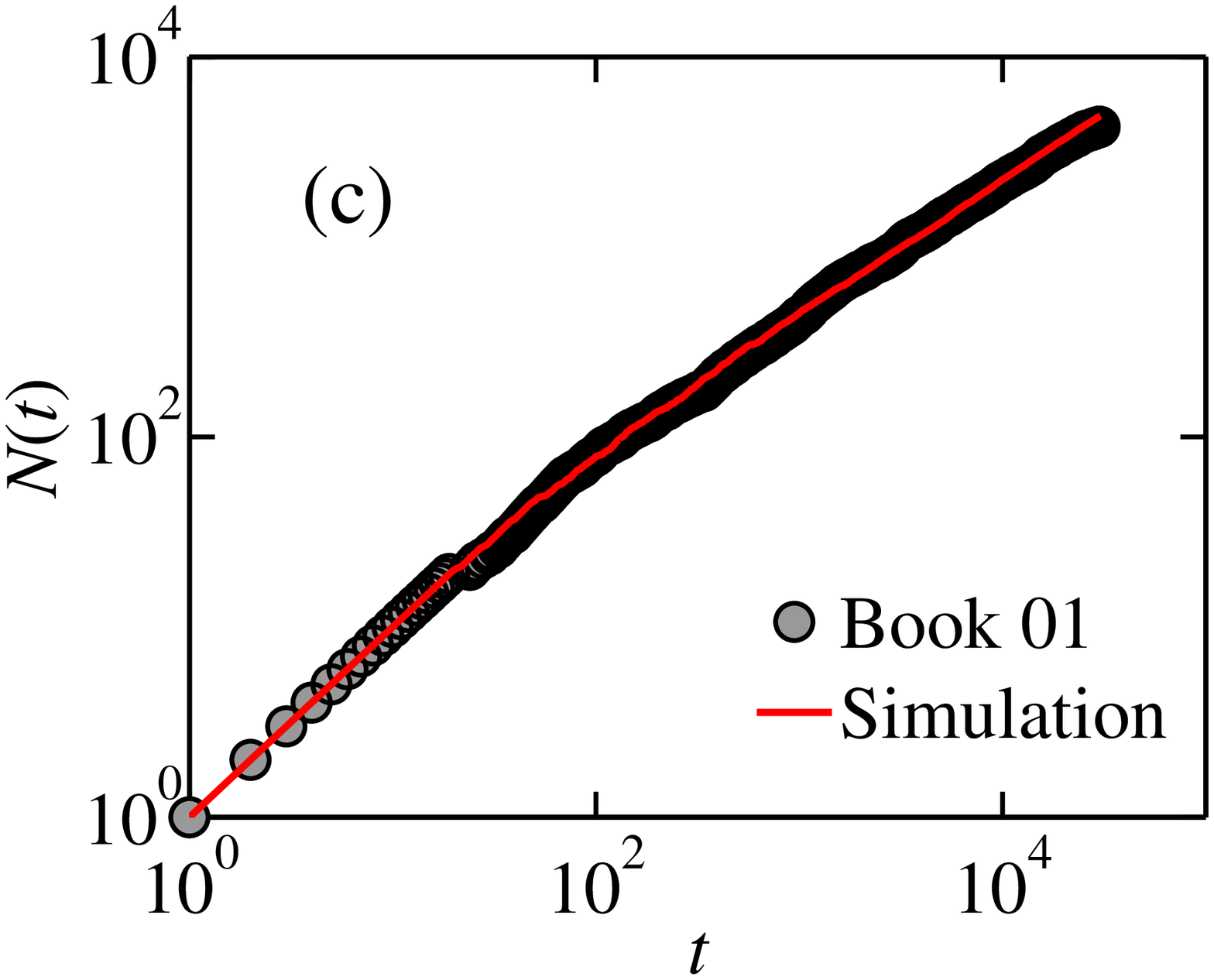}
}
\centerline{
\includegraphics[width=0.33\linewidth]{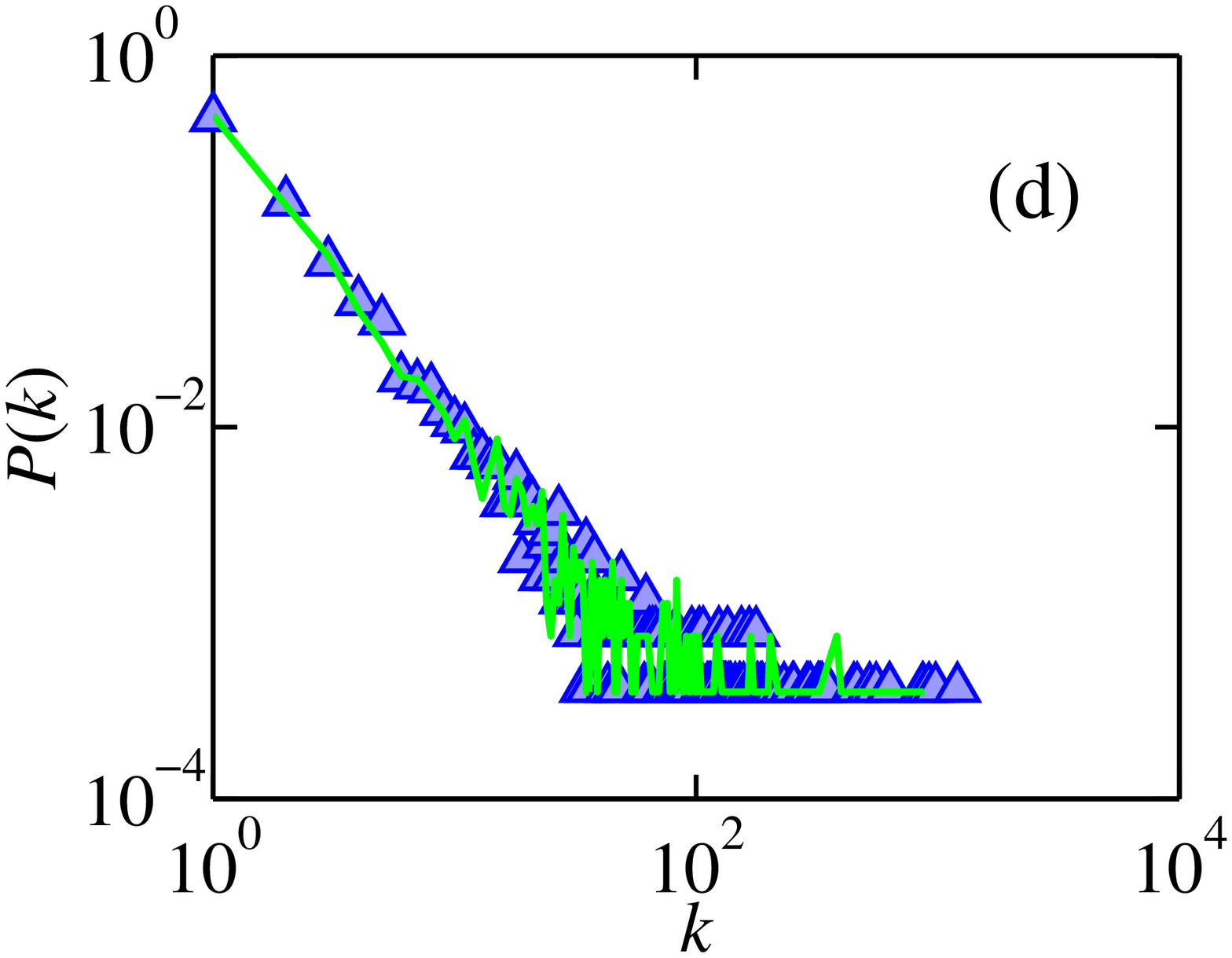}
\includegraphics[width=0.33\linewidth]{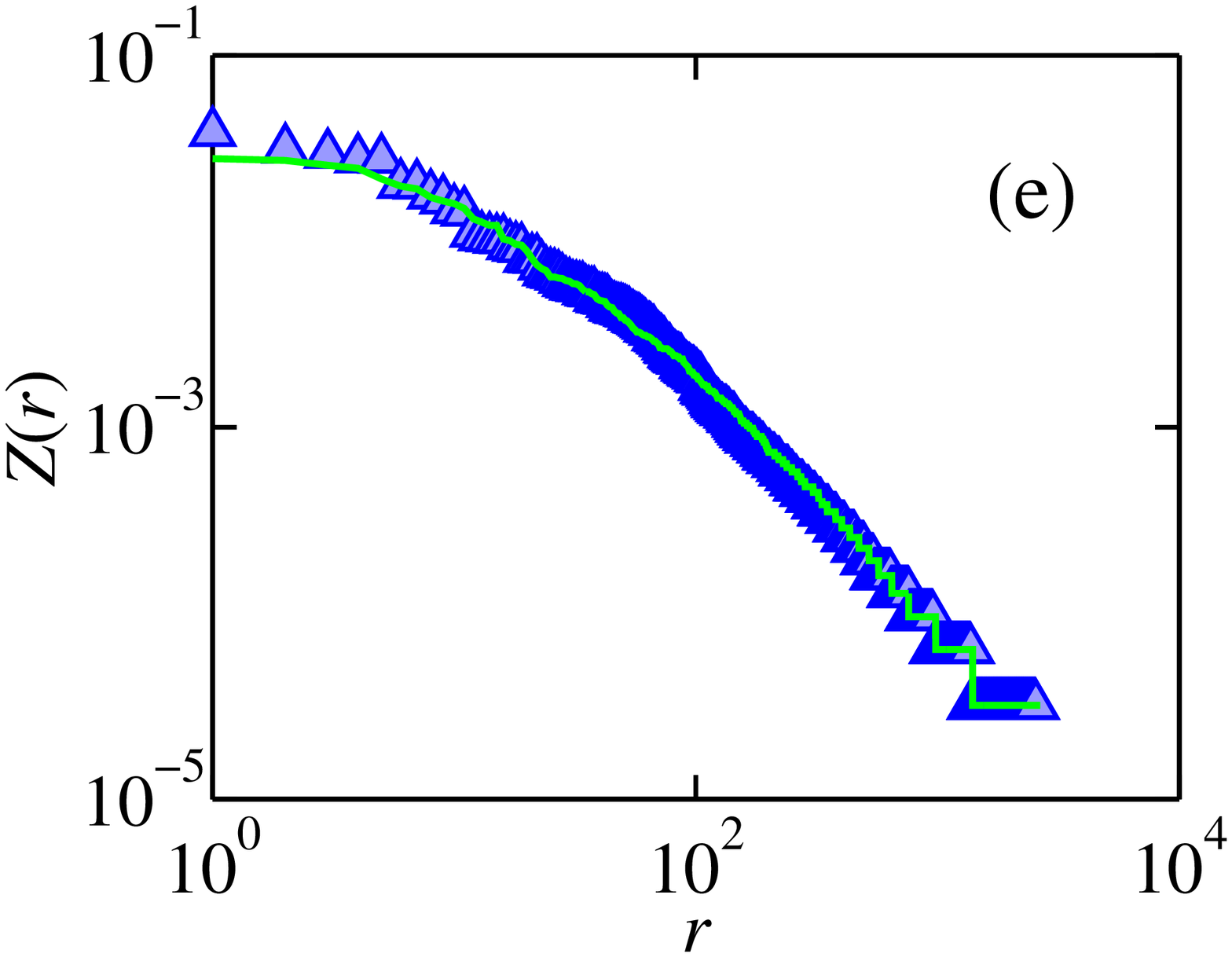}
\includegraphics[width=0.33\linewidth]{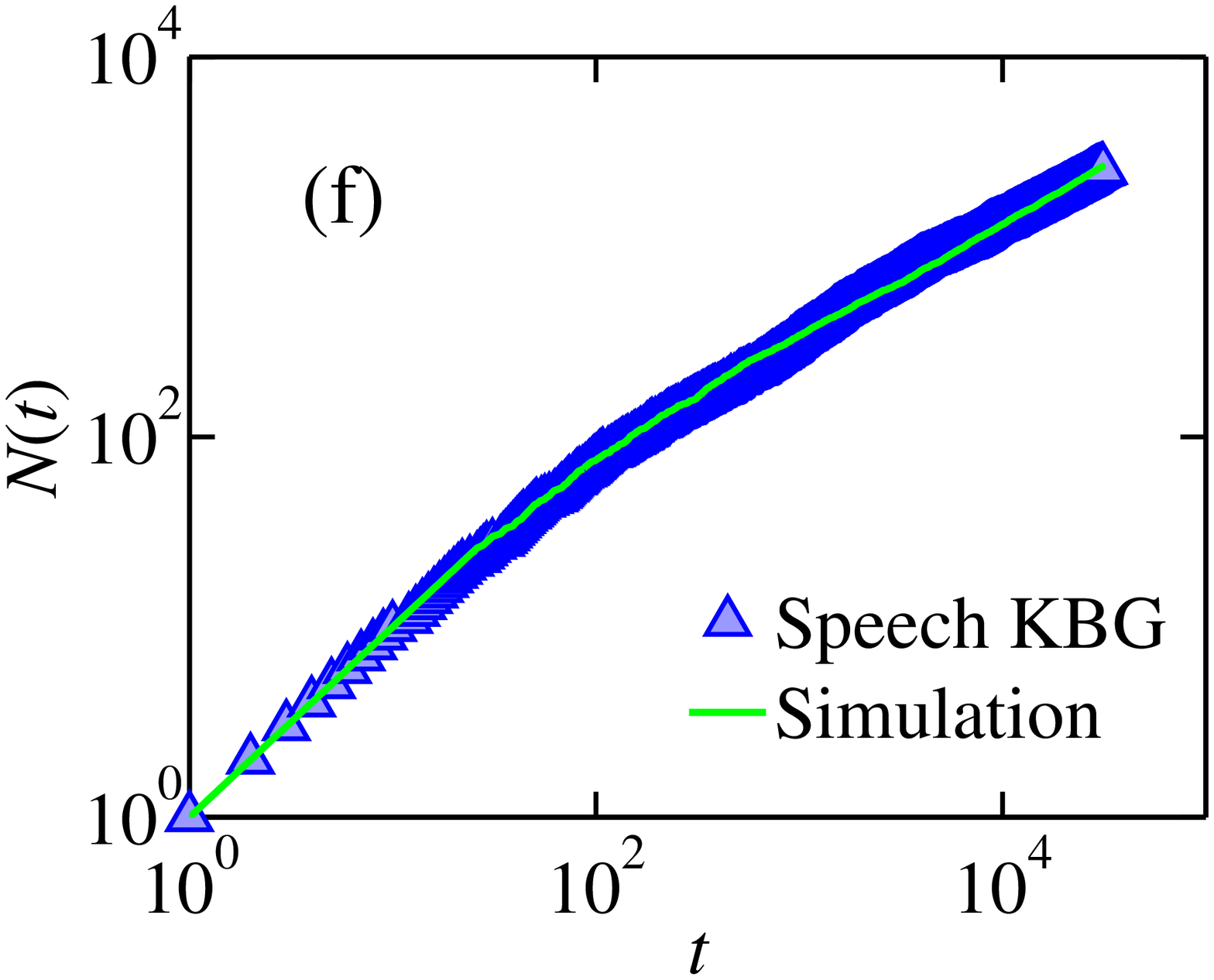}
}
\caption{(a), (b), (c) are the simulation results and the real data for books , and (d),(e),(f) are the  simulation result and the real data for speech. Modelling parameters for Book 01 are set as $k_0=2.34$, $k_t=0.29$ and $k_p=1.14$; and for Speech KBG $k_0=3.31$, $k_t=0.40$ and $k_p=1.08$.}
\end{figure}

\section*{Discussion}
 
Previous statistical analyses about human languages mostly concentrated on written texts where language consists of a huge number of words. In contrast, speech languages consisting of fewer words received less attention. The empirical results of the comparison between books and conversations indicate that i) the speech transcriptions also obey Zip’s law and Heaps' law, but with different exponents compared with books. ii) when the content length of speech grows, the emergence of new words does not increase as much as in books, the Heaps' law is also deviated from linear behavior. iii) in speech, the usage of words are much more concentrated on some words, which leads to the larger probability of high frequency rank words than in books. We can further explain the possible reasons. i) Book authors could create new words according to his own written style. New words may result from new techniques, new biological species, or new names. However, generally we seldom give birth to a new word during speech. ii) Book novel needs personalized language to express the characteristics of different persons in the story. Using complex grammatical long sentences to express the thoughts. Changing the style of words expression with different scenarios. Take Hamlet as an example, Claudius's speech is rich with rhetorical figures, as is Hamlet's and, at times, Ophelia's，while the language of Horatio, the guards, and the grave diggers is simpler.  While a great speech puts the occasion, the audience, and the speaker together in an unforgettable way. Most estimated the number of words per minute around $80 \sim 150$. So the speaker always choose a more simple way to use words.

The currently reported regularities from the well-known Zipf's and Heaps' laws point of view, can be reproduced by considering decrease of new words emergence with the length of generated text in a nonlinear process, which can be formulated according to the specific growing dynamics under study. In addition, the differences we observed in word frequency distribution and Zipf's law between speech and books indicate different strength of preferential attachment, which is also considered in the presented model. Simulation results confirm that the scaling properties of the complex dynamics of language construction and organization can be very well captured by the proposed model, and the differences between spoken language and written language can also be accounted for by different parameter settings. We further hypothesize that the proposed model of construction and organization in complex system with preferential attachment mechanism can also be applied to other complex system in physical, biological and social areas.

\section*{Methods}
\noindent \textbf{Data description} 10 books are analyzed in this article: No.1, Alice’s adventures in wonderland, written by Lewis Carroll; No.2, The adventures of Tom Sawyer, written by Mark Twain; No.3, A Christmas carol, written by Charles Dickens; No.4, David Crockett, written by John S. C. Abbott; No.5, An enquiry concerning human understanding, written by David Hume; No.6, Hamlet, written by William Shakespeare; No.7, The hound of the Baskervilles, written by Sir Arthur Conan Doyle; No.8, Moby-Dick: or, the whale, written by Herman Melville; No.9, The origin of species by means of natural selection, written by Charles Darwin; No.10, Ulysses, written by James Joyce. These books cover disparate topics and types and were accomplished in far different dates. The basic statistics of these books are presented in Table 1. This corpus of all the ten Chinese books consisting of about 1,002,504 total words and 100,335 distinct words.

The transcriptions of the Speech British National Corpus from Phonetics Laboratory are used in This article: The British National Corpus (BNC) is a 100 million word collection of samples of written and speech language from a wide range of sources, designed to represent a wide cross-section of British English from the later part of the 20th century, both speech and written. The speech part consists of orthographic transcriptions of unscripted informal conversations (recorded by volunteers selected from different age, region and social classes in a demographically balanced way) and speech language collected in different contexts, ranging from formal business or government meetings to radio shows and phone-ins. We selected 10 transcriptions of daily conversation from BNC database which have comparable length as the books. This corpus of all the ten speech text consisting of about 747,454 total words and 43,468 distinct words.

\noindent \textbf{Goodness of Linear Fit Statistics in Log-log Scale} The original data $P(k)$, $Z(r)$ and $N(t)$ are calculated in linear intervals, in log-log scale a lot of data cluster at large scales (see Fig. S1 for the data clustering at large scales). If the fitting process is performed on the original data, the data points at large scales will dominant the cost function and the measure of goodness of fit, causing bias towards large scales when fitting data in log-log scale. Thus, before the fitting process, we resample the original data to be equally distributed in log scale, as

\begin{equation}
y'(i) = \frac{\sum_{j=b^i}^{b^{i+1}} y_{\mathrm{orig}}(j)} {b^{i+1}-b^i},~~~i=1,2,...,\lfloor \log_bN \rfloor,
\label{equ:fit1}
\end{equation}

\noindent where $N$ is the number of the original data, and b can be any real number that greater than 1.

Then, we fit the resampled data in a least-squares sense in log-log scale, as to minimize the following cost function

\begin{equation}
SSE = \sum_{i=1}^{n} [\log(y'(i))-\log(y'_{\mathrm{fit}}(i))]^2,
\label{equ:fit2}
\end{equation}

\noindent and the statistic measures of the goodness of fit in log-log scale is defined following the definition of R-square as

\begin{equation}
R^2 = 1 - \frac{\sum_{i=1}^{n} [\log(y'(i))-\log(y'_{\mathrm{fit}}(i))]^2} {\sum_{i=1}^{n} [\log(y'(i))-\mathrm{E}(\log(y'(i)))]^2},
\label{equ:fit3}
\end{equation}

\noindent where E($\bullet$) denotes the calculation of mean value, and n is the number of the resampled data.

\noindent \textbf{Preferential attachment.} 
For each speech or book text, we divide it into two parts: Part I contains a fraction $\rho$ of words appeared and Part II contains the remain fraction $1-\rho$ of words. For word $i$ in Part II, if $i$ appeared $k$ times in Part I, we add one to $\phi(k)$ whose initial value is zero. Accordingly, $\phi(k)$ is the number of words in Part II that appeared $k$ times in Part I. Dividing $\phi(k)$ by the number of distinct words that appeared $k$ times in Part I. In this paper, we show the results for $\rho=0.5$.

\section*{Acknowledgements}

We thank the support from the Fundamental Research Funds for the Central Universities of China，Nanjing university scholar exchange project, Natural Science Foundation of China (61271082, 61201029), Priority Academic Program Development of Jiangsu Higher Education Institutions (PAPD), and Natural Science Foundation of Jiangsu Province (BK2011759). 

\section*{Author contributions}

C.H.B conceived the research. R.K.L and Q.L.M analysed the data, C.H.B and Q.L.M designed and simulated the model. All authors wrote and revised the manuscript.

\section*{Competing financial interests}

The authors declare no competing financial interests.

\section*{References}
\noindent 1. Searls, D. B. The language of genes. {\it Nature} \textbf{420}, 211–217 (2002). 

\noindent 2. Furusawa, C. \& Kaneko, K. Zipf's Law in Gene Expression. {\it Phys. Rev. Lett.} \textbf{90}, 088102 (2003).

\noindent 3. Koonin, E. V., Wolf, Y. I. \& Karev, G. P. The structure of the protein universe and genome evolution. {\it Nature} \textbf{420}, 218–223 (2002)

\noindent 4. Pigolotti, S., Flammini, A., Marsili, M. \& Martian, A. Species lifetime distribution for simple models of ecologies. {\it Proc. Natl. Acad. Sci. U.S.A.} \textbf{102}, 15747–15751 (2005). 

\noindent 5. Simon, H. A. \& Bonini, C. P. The size distribution of business firms. {\it Am. Econ. Rev.} \textbf{48}, 607–617 (1958).

\noindent 6. Cattuto C. Semiotic dynamics in online social communities. {\it Eur. Phys. J. C} \textbf{46}, 33–37 (2006). 

\noindent 7. Gabaix, X., Gopikrishnan, P., Plerou, V. \& Stanley, H. E., A theory of power-law distributions in financial market fluctuations. {\it Nature} \textbf{423}, 267–270 (2003).

\noindent 8. Shao, J., Ivanov, P. Ch., Urosevic, B., Stanley, H. E. \& Podobnik B., Zipf rank approach and cross-country convergence of incomes. {\it Europhys. Lett.} \textbf{94}, 48001 (2011).

\noindent 9. Brakman, S., Garretsen, H., van Marrewijk C. \& van den Berg, M. The return of Zipf: Towards a further understanding of the rank-size distribution. {\it J. Reg. Sci.} \textbf{39}, 739–767 (1999).

\noindent 10. Gabaix, X. Zipf's law for cities: an explanation. {\it Q J Econ} \textbf{114}, 739–767(1999).

\noindent 11. Nowak, M. A. \& Krakauer, D. C. The evolution of language. {\it Proc. Natl. Acad. Sci. U.S.A.} \textbf{96}, 8028–8033 (1999).

\noindent 12. Dorogovtsev, S. N. \& Mendes, J. F. F. Language as an evolving word web. {\it Proc. R. Soc. Lond. B} \textbf{268}, 2603–2606 (2001).

\noindent 13. Nowak, M. A., Komarova, N. L. \& Niyogi, P. Computational and evolutionary aspects of language. {\it Nature} \textbf{417}, 611–617 (2002).

\noindent 14. Hauser, M. D., Chomsky, N. \& Fitch, W. T. The faculty of language: what is it, who has it, and how did it evolve? {\it Science} \textbf{298}, 1569–1579 (2002).

\noindent 15. Abrams, D. M. \& Strogatz, S. H. Modelling the dynamics of language death. {\it Nature} \textbf{424}, 900 (2003).

\noindent 16. Lieberman, E., Michel, J. B., Jackson, J., Tang, T. \& Nowak, M. A. Quantifying the evolutionary dynamics of language. {\it Nature} \textbf{449}, 713–716 (2007).

\noindent 17. Sole, R. V., Corominas-Murtra, B., Valverde, S. \& Steels, L. Language networks: Their structure, function, and evolution. {\it Complexity} \textbf{15}, 20–26 (2010).

\noindent 18. Petersen, A. M., Tenenbaum, J., Havlin, S. \& Stanley, H. E. Statistical laws governing fluctuations in word use from word birth to word death. {\it Sci. Rep.} \textbf{2}, 313 (2012).

\noindent 19. Gao, J., Hu, J., Mao, X. \& Perc, M. Culturomics meets random fractal theory: insights into long-range correlations of social and natural phenomena over the past two centuries. {\it J. R. Soc. Interface} \textbf{9}, 1956–1964 (2012).

\noindent 20. Zipf, G. K. Behavior and the principal of least effort (Addison-Wealey,Cambridge, MA, 1949).

\noindent 21. Heaps, H. S. Information retrieval-computational and theoretical aspects (Academic Press, 1978).

\noindent 22. Kanter, I. \& Kessler, D. A. Markov processes: linguistics and Zipf's law. {\it Phys. Rev. Lett.} \textbf{74}, 4559–4562 (1995).

\noindent 23. Cancho, R. F. i. \& Sole, R. V. Least effort and the origins of scaling in human language. {\it Proc. Natl. Acad. Sci. U.S.A.} \textbf{100}, 788–791 (2002).

\noindent 24. Gelbukh, A. \& Sidorov, G. Zipf and Heaps Laws coefficients depend on language. {\it Lect. Notes Comput. Sci.} \textbf{2004}, 332–335 (2001).

\noindent 25. Serrano, M. A., Flammini, A. \& Menczer, F. Modeling statistical properties of written text. {\it PLoS ONE} \textbf{4}, e5372 (2009).

\noindent 26. Wang, D., Li, M. \& Di, Z. True reason for Zipf's law in language. {\it Physica A} \textbf{358}, 545 (2005).

\noindent 27. Lu, L., Zhang, Z. \& Zhou, T. Deviation of Zipf's and Heaps' laws in human languages with limited dictionary sizes. {\it Sci. Rep.} \textbf{3}, 1082 (2013).

\noindent 28. Cattuto, C., Loreto, V. \& Pietronero, L. Semiotic dynamics and collaborative tagging. {\it Proc. Natl. Acad. Sci. U.S.A.} \textbf{104}, 1461–1464 (2007).

\noindent 29. Cattuto, C., Barrat, A., Baldassarri, A., Schehr, G. \& Loreto, V. Collective dynamics of social annotation. {\it Proc. Natl. Acad. Sci. U.S.A.} \textbf{106}, 10511–10515 (2009).

\noindent 30. Zhang, Z.-K., Lu, L., Liu, J.-G. \& Zhou, T. Empirical analysis on a keyword-based semantic system. {\it Eur. Phys. J. B} \textbf{66}, 557–561 (2008).

\noindent 31. Lansey, J. C. \& Bukiet, B. Internet search result probabilities: Heaps' law and word associativity. {\it J. Quant. Linguistics} \textbf{16}, 40–66 (2009).

\noindent 32. Zhang, H.-Y. Discovering power laws in computer programs. {\it Inf. Process. Manage.} \textbf{45}, 477–483 (2009).

\noindent 33. Coleman, J. ,Baghai-Ravary, L., Pybus, J., \& Grau, S. Audio BNC: the audio edition of the Speech British National Corpus. (Phonetics Laboratory, University of Oxford 2012). 

\noindent 34. Clauset, A. \& Moore, C. Accuracy and scaling phenomena in internet mapping. {\it Phys. Rev. Lett.} \textbf{94}, 018701 (2005).
%Clauset, A., Shalizi, C. R. \& Newman, M. E. J. Power-law distributions in empirical data. {\it SIAM Rev.} \textbf{51}, 661–703 (2009).

\noindent 35. Jeong, H., Neda, Z. \& Barabasi, A.-L. Measuring preferential attachment for evolving networks. {\it Europhys. Lett.} \textbf{61}, 567–572 (2003).

\noindent 36. Barabasi, A.-L. \& Albert, R. Emergence of scaling in random networks. {\it Science} \textbf{286}, 509–512. (1999).

\noindent 37. Dorogovtsev, S. N. \& Mendes, J. F. F. Effect of the accelerating growth of communications networks on their structure. {\it Phys. Rev. E} \textbf{63}, 025101(R) (2001).

\noindent 38. Yule, G. U. A mathematical theory of evolution, based on the conclusions of Dr. J. C. Willis, F.R.S. {\it Phil. Trans. R. Soc. B} \textbf{213}, 21–87 (1925).

\noindent 39. Simon, H. A., On a class of skew distribution functions. {\it Biometrika} \textbf{42}, 425–440 (1955).

\noindent 40. Champernowne, D. A model of income distribution. {\it Econ J.} \textbf{63}, 318–351 (1953).

\noindent 41. Chung K. H., \& Cox R. A. K. A stochastic model of superstardom: An application of the Yule distribution. {\it Rev Econ Stat} \textbf{76}, 771–775 (1994).

\noindent 42. Cattuto, C., Loreto, V. \& Servedio, V. D. P. A Yule-Simon process with memory. {\it Europhys. Lett.} \textbf{76(2)}, 208–214 (2006).

\noindent 43. Garibaldi, U., Costantini, D., Donadio, S. \& Viarengo, P. Herding and clustering in economics: The Yule-Zipf-Simon model. {\it Computational Econ.} \textbf{7}, 115–134 (2006).

%\noindent 40. Cattuto, C., Loreto, V. \& Pietronero, L. Semiotic dynamics and collaborative tagging. {\it Proc. Natl. Acad. Sci. U.S.A.} \textbf{104}, 1461–1464 (2007).

\bigskip
\bigskip
\bigskip

%\clearpage

\end{document}